\documentclass[twocolumn]{bmcart}

\usepackage{amsthm,amsmath,amssymb}
\usepackage{xcolor,url,tabularx,multirow,pgfgantt,booktabs,graphicx,hyperref}
\usepackage{dblfloatfix}
\usepackage{soul}
\usepackage{url}

\RequirePackage{natbib}
\RequirePackage{hyperref}
\usepackage[utf8]{inputenc} 
\usepackage{amstext}

\newcolumntype{C}[1]{>{\centering\arraybackslash}m{#1}}

\definecolor{answer}{rgb}{0,0,0.6}
\definecolor{todo}{rgb}{0,0.55,0}
\definecolor{open}{rgb}{0.9,0,0}

\colorlet{mygreen}{green!60!gray}

\newcommand{\CH}[1]{\textcolor{black}{{#1}}}

\begin{document}

\begin{frontmatter}

\begin{fmbox}
\dochead{Research}


\title
{Boosting CNN-based primary quantization matrix estimation of double JPEG images via a classification-like architecture}


\author[
   addressref={aff1},                   
   corref={aff1},                       
   email={benedettatondi@gmail.com}   
]{\inits{BT}\fnm{Benedetta} \snm{Tondi}}
\author[
   addressref={aff1},
   email={andreacos82@gmail.com}
]{\inits{AC}\fnm{Andrea} \snm{Costanzo}}
\author[
   addressref={aff2},
   email={libin@szu.edu.cn}
]{\inits{DH}\fnm{Dequ} \snm{Huang}}
\author[
   addressref={aff2},
   email={libin@szu.edu.cn}
]{\inits{BL}\fnm{Bin} \snm{Li}}



\address[id=aff1]{
  \orgname{Department of Information Engineering and Mathematics, University of Siena}, 
  \street{via Roma 56},                     %
  \postcode{53100}                                
  \city{Siena},                              
  \cny{Italy}                                    
}
\address[id=aff2]{%
  \orgname{Guangdong Key Laboratory of Intelligent Information Processing and Shenzhen Key Laboratory
of Media Security, Shenzhen University},
  \street{No 3688 Nanhai Avenue},
  \postcode{518060}
  \city{Shenzhen},
  \cny{China}
}





\begin{abstractbox}

\begin{abstract} 
%
\CH{Estimating the  primary quantization matrix of double JPEG compressed images
is a problem of relevant importance in image forensics
since it allows to infer
important information about the past history of an
image. In addition, the inconsistencies of the primary quantization matrices across different image regions can be used to localize
splicing in double JPEG tampered images.
Traditional model-based approaches work under specific assumptions on the relationship between the first and second compression qualities and on the alignment of the JPEG grid.
Recently, a deep learning-based estimator
capable to work under a wide variety of conditions  has been proposed, that  outperforms tailored existing methods  in most of the cases.
The method is based  on a Convolutional Neural Network (CNN) that is trained to solve the estimation as a standard regression problem.%
By exploiting the integer nature of the quantization coefficients, in this paper, we propose a deep learning technique that performs the estimation by resorting to a simil-classification architecture. The CNN is  trained
with a loss function that takes into account both the accuracy and the Mean Square Error (MSE) of the estimation.
Results confirm the superior performance of the proposed technique, compared to the state-of-the art methods based on statistical analysis and, in particular, deep learning regression.
Moreover, the capability of the method to work under general operative conditions, regarding the alignment of the second compression grid with the one of first compression and the combinations of the JPEG qualities of former and second compression,
is very relevant in practical applications, where these information are unknown a priori.}
\end{abstract}




\end{abstractbox}
\end{fmbox}

\end{frontmatter}





\section{Introduction}


Detection of double JPEG (DJPEG) compression is one of the most widely studied problems in image forensics, see for instance \cite{Jessica2008,Li2008,Barni2017}. The interest of researcher in this topic is motivated by the fact that double compression can reveal important information about the past history of an image.
Important information can be obtained by estimating the quality of the first JPEG compression, and, moreover, by estimating the primary quantization matrix used for the first JPEG compression. Given an image with several  copy-pasted regions, it is possible to identify the different origin of the tampering by recognizing that they have been compressed first with different JPEG qualities, and more in general that they are characterized by different primary quantization matrices of the compression  (while there is not a standard definition of the JPEG quality, the concept of quantization matrix is a standard one \cite{pennebaker1992}).

Several methods have been proposed in the literature for the estimation of the primary quantization matrix. Many of them exploit statistical modeling of DCT coefficients \cite{Bianchi2012,Cogranne2019,Galvan2014,Dalmia2018}.
A common feature of all these approaches is that they work under particular operative conditions and settings about the relationship
of the JPEG qualities of former and second compression, and the alignment of the $8 \times 8$ grid of the first compression with the second one.
For instance, the method in \cite{Galvan2014} works only when the two compressions are aligned and
the second quantization step is lower than the first one,
that is when the quality of the second JPEG is higher than the quality of the first one (hence, $QF_{1} < QF_{2}$ for the standard quantization matrix case). Similarly, the method in \cite{Cogranne2019} is designed for the aligned JPEG case and can not estimate the first quantization step when this is a divisor of the second one.
\CH{A more general approach for the estimation in the aligned scenario has been recently proposed in \cite{battiato2020indepth}.}
The algorithm proposed in \cite{Bianchi2012} can work both in the aligned and non-aligned cases, however the performance drops when $QF_{1} > QF_{2}$. Eventually, the method in \cite{Dalmia2018} works in the non-aligned case only.
Another drawback of such model-based techniques and approaches that rely on hand-crafted features
is that their performance tend to decrease significantly when they are applied to small patches, that prevents the application of these methods for the local estimation of the quantization matrix of first compression, useful for tampering localization.

A modern method for primary quantization matrix estimation based on Convolutional Neural Networks (CNN) has been recently proposed in \cite{niu2019SPL}. Such method can work under very general operative conditions and on small (64$\times$64) patches. This approach has been shown to outperform  previous approaches,
both in terms of accuracy and mean square error (MSE) of the estimation.
In particular, in \cite{niu2019SPL}, the CNN is trained to minimize the squared difference between the predicted values of the quantization coefficients and the true values, hence the MSE of the estimation is minimized.
%
%
%
Some works in the deep learning literature, however, shows that CNNs are better to solve classification than regression problems.
CNNs can in fact achieve remarkably accurate results when trained to predict categorical variables, drawn from discrete probability distributions of data \cite{bulat2016human,chen2017deeplab}.
%
Whenever possible, switching to a classification problem or consider hybrid methods that combine classification with regression has been shown to yield better results \cite{alp2017densereg}.
In \cite{alp2017densereg}, for instance,
soft values are estimated  by using a quantized regression architecture that first obtains a quantized estimate (using the softmax followed by the cross entropy loss), and then refines it through regression of the residual.

Given the above, in this paper, we focus on improving the performance of CNN-based estimation of the primary quantization matrix by turning the regression into a classification-like problem, with the design of a suitable CNN architecture.
Our approach starts from the observation that the quantization coefficients can only take integer values.
%
%
Therefore, we design a structure such that the estimation of a vector of integer values, namely all the coefficients of the quantization matrix, can be performed in a classification-like fashion.
For the implementation of the network (internal layers) we consider the same CNN architecture already considered in \cite{niu2019SPL}, yielding good results, namely DenseNet \cite{huang2017densely}.
Similarly to \cite{niu2019SPL}, the CNN-based estimator
is designed to work under very general operative conditions, i.e. when the second compression grid is either aligned or not with the one of first compression, and for every combinations of qualities of former and second compression.
%
%
%
%
The capability of the method
to work  under both aligned
and non-aligned DJPEG, and for all possible combinations of
JPEG qualities, is very relevant in
practical applications, where those information
are not known a priori, thus making the adoption of dedicated method very impractical.
%
%
%
%
\CH{Like in \cite{niu2019SPL}, we focus on the case where the estimation is carried out on small patches, that represents the most
challenging scenario.}
%
%
%
%
\CH{As commonly done by the approaches from the literature of primary quantization matrix estimation,
we assume that the test image is double compressed and do not consider the single JPEG scenario (in this case, quite reasonably,
our method returns a quantization matrix that corresponds to a very high compression quality, the estimated coefficients being close to those of the quantization matrix for the case $QF=100$. Said differently, the network regards the single compression with $QF$
as a double compression with $QF_1 = 100$ and $QF_2 = QF$).}

The rest of this paper is organized as follows: Section \ref{sec.background} recaps the main concepts of double compression and introduces the notation. The proposed method is described in Section \ref{sec.prop_method}. Then, Section \ref{sec.exp_method} details the experimental
methodology and the results are reported  and discussed in Section \ref{sec.results}. We conclude the paper with some final remarks in Section \ref{sec.conclusions}.

\section{Basic concepts and notation}
\label{sec.background}


We denote by $Q$ the quantization matrix, that is, the $8 \times 8$ matrix with the quantization steps of the DCT coefficients considered for the compression.
A double compression occurs when an image compressed with a given  $Q_1$ is decompressed (decompression involves de-quantization and inverse DCT), and compressed again with a second quantization matrix $Q_2$.
%
%
The elements of $Q_1$ can be conveniently arranged in a vector of dimensionality 64, zig-zag ordered \cite{pennebaker1992}. We denote by ${\bf q}_1$ such 64-dim vector built from $Q_1$.
As commonly done in the literature \cite{Bianchi2012,Cogranne2019,Galvan2014, niu2019SPL}, we focus on the first elements of ${\bf q}_1$ and restrict the estimation to those coefficients.
We denote  with $({\bf q}_{1})_{N_c} = [q_{1,1}, q_{1,2},...,q_{1,N_c}]$  the vector of the first $N_c$ coefficients of ${\bf q}_1$.
The coefficients at the medium-high DCT frequencies are in fact more difficult to estimate accurately, due to the stronger quantization usually applied to them; however, since these coefficients are not very discriminative (as they tend to be similar for most quantization matrices), their estimation is less important.
%
%

When a JPEG image is compressed a second time, the second compression grid can be either aligned or non-aligned to the first compression grid.
%
%
The case of a non-aligned DJPEG corresponds to the most frequent scenario in practice. A grid misalignment
occurs locally when image splicing is performed, that is, when a region of a single JPEG image is copy-pasted into another image, since in this case the alignment between the compression grids is rarely preserved. 
On a global level, we have a non-aligned DJPEG  when the image is cropped in between the former and second compression stage, or some processing is applied causing a de-synchronization.

The quality of the JPEG compression is often summarized by many compression softwares by means of the JPEG Quality Factor ($QF$),  whose values range from 0 to 100 ($QF$ values lower than 50 however are seldom used in practice nowadays since they corresponds to extremely low qualities). A $QF$ value specifies a quantization matrix $Q$ (standard quantization matrix).
%
For convenience, in the rest of the paper,  we refer to the JPEG Quality Factor ($QF$).
Note that,
in principle, the proposed estimator can be applied to estimate any quantization matrix of former compression, be it standard and non-standard.
In the rest of the paper, we denote with $QF_2$ the second compression $QF$ and with $QF_1$ the former.

\CH{Like in \cite{niu2019SPL}, in this paper, we consider color DJPEG images (with 3 channels) and focus on the estimation of the primary quantization matrix $Q_1$ of the luminance channel, that is,  the $Y$ channel of the $Y C_b C_r$ color space \cite{pennebaker1992}.}

\section{Proposed classification-like  CNN estimator}
\label{sec.prop_method}

%

The proposed method starts from the observation that the
$q_{1,i}$'s values are discrete values.
In \cite{niu2019SPL}, where a regression problem is addressed to estimate $({\bf q}_{1})_{N_c}$, the values obtained at the output of the CNN are finally quantized to get the estimated vector $(\hat{{\bf q}}_{1})_{N_c}$ (specifically, rounding is performed on each element of the output vector independently, yielding $\hat{q}_{1,i}$, $i = 1,...,N_c$).
%
%
However, it has been shown that
for the estimation of discrete quantities (following then a categorical distributions)  it is often preferable to resort to softmax followed by the cross entropy loss, which is good for backpropagation  (see \cite{alp2017densereg}).
%
Therefore, we propose to
switch the regression to a classification-like problem. To do so, we consider
a custom output layers structure
with a basic loss function, and also with a refined loss function, described in the following.
%
%

\subsection{General structure}


The architecture that we considered for the internal layers of the CNN, and in particular, the feature extraction part, is a dense structure,  namely the DenseNet backbone architecture  \cite{huang2017densely}, and is described in Section \ref{sec.architecture}.
In the following, we describe the specific structure of the proposed output layer.

In the proposed structure, each to-be-estimated coefficient ${q}_{1,i}$ ($i=1,2,..,N_c$) of discrete value is encoded
as a one-hot vector.
The dimensionality of the encoded vector is determined by all the possible values that ${q}_{1,i}$  may take. Assuming that the quality of the image can not be too low, in fact, for every $i$,
%
the estimated coefficient $\hat{q}_{1,i}$ may take a limited number of values, that is, $1 \le \hat{q}_{1,i} \le q_{1,i}^M$. For simplicity,
we set $q_{1,i}^M$ equal to the corresponding value of the $i$-th coefficient when $QF_1 = 50$ (minimum quality of the JPEG considered).\footnote{This corresponds to assume that the former JPEG quality is always higher than or equal to $QF_1 = 50$, which is often the case in practice, thus not representing a big restriction (as a consequence, $Q_1$ matrices corresponding to lower qualities are not correctly estimated).}
To get the desired output, we set the logit level output to a size $[q_{1,1}^M + q_{1,2}^M \cdots + q_{1,N_c}^M]$; then,
the softmax is applied block-based on each $N_c$ block, where each block has  $q_{1,i}^M$ inputs, $i=1,2,..,N_c$.
%


\begin{figure*}[t!]
\begin{center}
\includegraphics{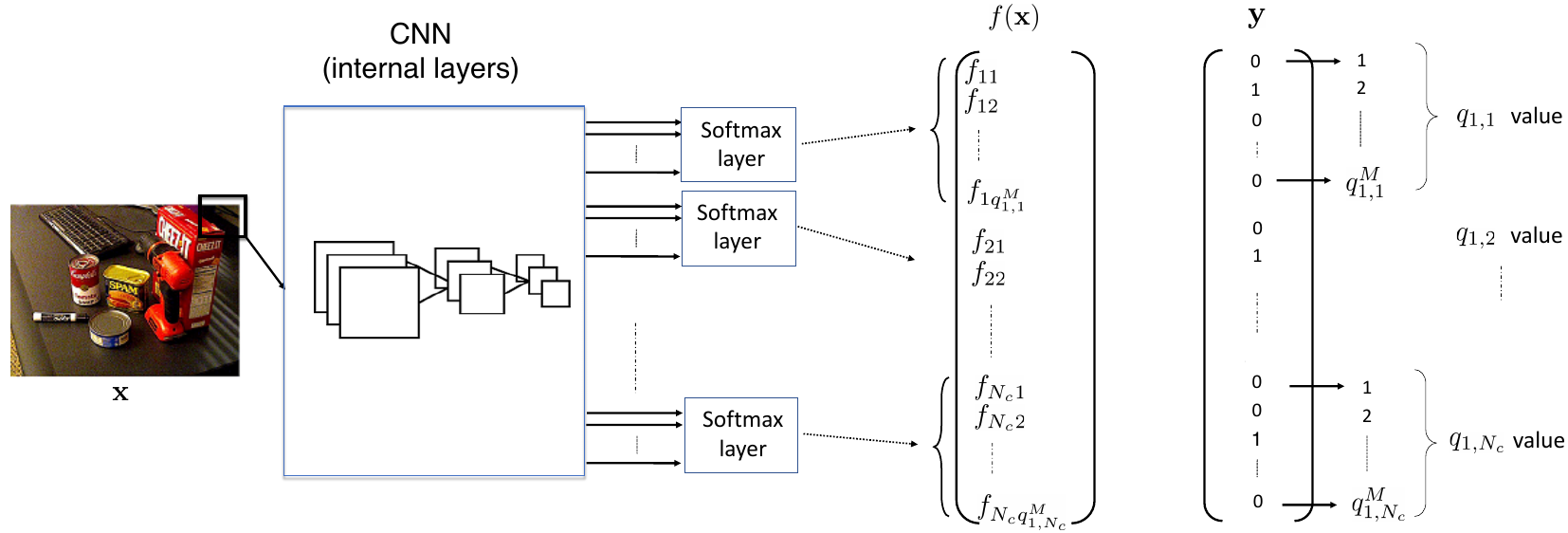}
\caption{Scheme of the CNN classification-like structure considered in this paper. }
\label{fig:setup}
\end{center}
\end{figure*}

For training the network we consider the following basic custom loss:
\begin{equation}
\mathcal{L}({\bf x}) =  \frac{1}{N_c} \sum_{i=1}^{N_c} \left(\sum_{j \in [1: q_{1,i}^M]} y_{ij} \log(f_{ij}({\bf x})) \right),
\label{eq.loss}
\end{equation}
where ${y}_{i,j}$ denotes the ground-truth label corresponding to ${q}_{1,i}$.
According to Eq. \eqref{eq.loss}, a cross-entropy loss is first computed on each block separately, then the
loss is defined as the sum of all the cross-entropy loss terms.

Figure \ref{fig:setup} illustrates the scheme of the CNN considered for the estimation with specific focus on the output layer.
In the figure, ${\bf y}$ denotes the ground-truth vector of the $N_c$ one-hot encoded vectors, having the dimensionality $[q_{1,1}^M + q_{1,2}^M + \cdots  + q_{1,N_c}^M]$, and ${f}({\bf x})$ the output soft vector of the CNN, having the same dimensionality of ${\bf y}$.
Formally, ${\bf y} = {\bf y}_1 \oplus {\bf y}_2 \oplus \cdots \oplus {\bf y}_{N_c}$, where $\oplus$ denotes the horizontal concatenation, that is,
\begin{equation}
{\bf y} = [y_{11}, y_{12},...,y_{1 q_{1,1}^M}, y_{21}, \cdots y_{N_c 1}, ..., y_{N_c q_{1,N_c}^M}],
\end{equation}
%
and, similarly, $f({\bf x}) = [f_1({\bf x}) \oplus \cdots f_{N_c}({\bf x})]$ where $f_i({\bf x}) = (f_{ij}({\bf x}))_{j=1}^{q_{1,1}^M}$.


As we said,  $q_{1,i}^M$, for every $i$,  is determined considering the value assumed by the $i$-th coefficient in the quantization matrix corresponding to the lowest $QF_1$ considered (that we set to $50$ in our experiments).
Then, the final estimated vector $(\hat{\bf q}_{1})_{N_c}$ is given by
%
\begin{equation}
\hat{q}_{1,i} = \arg\max_j f_{ij}({\bf x}), \quad i=1,...,N_c.
\end{equation}


\subsection{Refined loss function}


For a given image ${\bf x}$, and final predicted vector $(\hat{{\bf q}}_{1})_{N_c}$,
the accuracy of the estimation  is averaged over all the $N_c$ coefficients,
that is,  Accuracy$({\bf x}) = (1/{N_c})\sum_{i=1}^{N_c}\delta({q}_{1,i}({\bf x}),\hat{{q}}_{1,i}({\bf x}))$, where  $\delta$ is the Kronecker delta ($\delta(a,b) = 1$ if $a = b$, $0$ otherwise).
The Mean Square Error (MSE) of the estimation is given by
${\text{MSE}}({\bf x}) = (1/N_c)\sum_{i=1}^{N_c} |q_{1,i}({\bf x}) - \hat{q}_{1,i}({\bf x})|^2$.

\begin{figure}[t!]
\begin{center}
\includegraphics{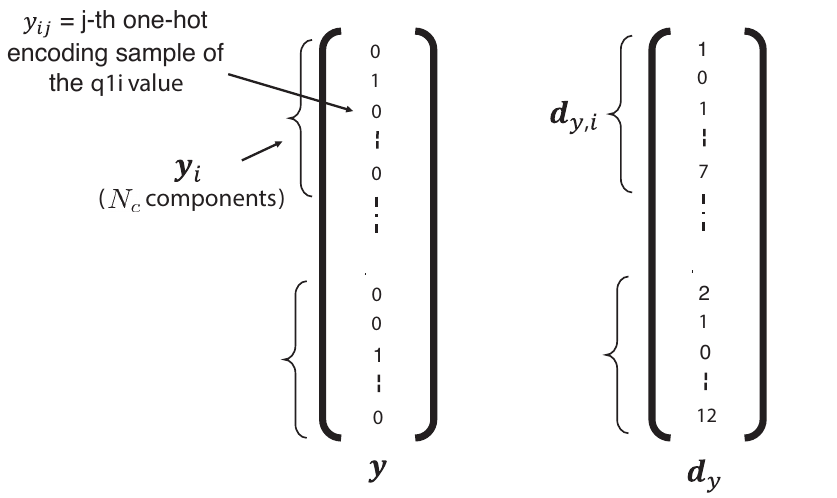}
\caption{Vector ${\bf d}_y$ of the relative distances obtained from ${\bf y}$.}
\label{fig:mapping-ytody}
\end{center}
\end{figure}

The new classification-like structure trained with the loss function $\mathcal{L}$ attempts to maximize the accuracy of the estimation,  without caring  about the MSE of the estimation. From  Eq.\eqref{eq.loss}, it is easy to argue that solutions yielding large MSE values are not penalized compared to those yielding lower values, for the same soft values associated to the '1' positions in vector $y$ ($\mathcal{L}({\bf x})$ takes the same value). Said differently, an incorrect decision on the value of a $q_{1,i}$ for some $i$, that results in a different wrong one-hot encoded vector, may lead to a same value of the loss function in Eq.\eqref{eq.loss}, regardless of the estimated value, or better yet, regardless of the difference between the true and estimated value, i.e., $|q_{1,i} - \hat{q}_{1,i}|$.
Since both the accuracy and the MSE of the estimation are important in practice, we would like to get high  accuracy for the estimation, without paying (much) in terms of MSE.
%
In order to solve this issue, we investigated two possible solutions, that corresponds to two possible refinements of the loss function.
The first solution was to use a ``smooth'' categorical cross-entropy loss that keeps all the advantages of the standard cross-entropy loss,
but at the same time assigns different weights to the errors depending on the position of the ``1'' inside each one-hot encoded vector, that is, depending on $|q_{1,i} - \hat{q}_{1,i}|$ for each $i$.
The second solution, that gave us to get better results, was to perform jointly classification and regression by considering a combined loss (as done by some approaches in the literature of  deep learning and standard machine learning \cite{joint1,joint2,joint3}).
%
%
Given the simil-classification architecture considered, defining a suitable loss function that takes into account the distance between the estimate and the true value,  and then penalizes large values of such distance,  is not obvious.
To do so, we define a vector  ${\bf d}_y$ that reports in each position the distance from the '1' in the corresponding one-hot encoded vector $y_{i}$ (see Figure \ref{fig:mapping-ytody}). Formally, let ${\bf d}_y = {\bf d}_{y,1} \oplus {\bf d}_{y,2} \oplus \cdots \oplus {\bf d}_{y,N_c}$.
 The $j$-th component of  ${\bf d}_{y,i}$, $i=1,\cdots N_c$,   is given by
%
\small
\begin{align}
& d_{y,i,j} =  \nonumber\\
& \hspace{-0.5cm} \left\{\begin{array}{ll}
%
%
|j - q_{1,1}| & 1 \le j \le q_{1,1}^M \\
|j - q_{1,1}^M - q_{1,2}| & q_{1,1}^M + 1 \le j \le q_{1,1}^M  + q_{1,2}^M\\
\cdots & \cdots\\
\big|j -   \left(\underset{i=1}{\overset{N_c - 1}{\sum}} q_{1,i}^M\right) - q_{1, N_c}\big| & \underset{i=1}{\overset{N_c - 1}{\sum}} q_{1,i}^M + 1 \le j \le  \underset{i=1}{\overset{N_c}{\sum}} q_{1,i}^M, \\
\end{array}\right.
\label{eq.dy}
\end{align}
\normalsize
where ${\bf q}_1$ is the true vector of coefficients.
%

Then, we define the combined loss function as follows:
%
%
\begin{align}
& \mathcal{L}^r({\bf x}) =  c \cdot \frac{1}{N_c}  \sum_{i=1}^{N_c}  \left(\sum_{j \in [1: q_{1,i}^M]}  y_{ij} \log(f_{ij}({\bf x})) \right) + \hspace{0.5cm} \nonumber\\
& \hspace{3cm}  + (1-c) \cdot \left(f^{T}({\bf x}) \cdot {\bf d}_y\right),
\label{eq.loss_comb}
\end{align}
%
where $c$ is a constant, $0 < c < 1$, determining the trade off between the two terms.


We observe that, for each $i$, the contribution to the second term is large when the $\arg\max_{j} f_{ij}({\bf x})$, that is $\hat{q}_{1,i}$, is far from $q_{1j}$, small otherwise (the second term is 0 when $\arg\max_{j} f_{ij}({\bf x}) = q_{1i}$ for every $i$, that is,  in the case of ideal estimation).
Then, the refined loss indirectly takes into account the MSE of the estimation via the second term.
Moreover, the second term is continuously differentiable
and then is good for backpropagation.

Some preliminary experiments we carried out confirmed that, as expected, adopting the loss function $\mathcal{L}^r$, instead of $\mathcal{L}$,  a significantly lower MSE can be reached, at the price of a possible slight decrease in the accuracy. Based on our experiments, by training our CNN model on a mixture of $QF_1$ ( $QF_1 = \{60,65,70,75,80,85,90,95,98\}$) with $QF_2 = 90$ for the same number of epochs (100) with the $\mathcal{L}$ and $\mathcal{L}^r$ loss,  we got an average accuracy and average MSE of 0.8511 and 1.990  in the first case
and 0.8516 and $0.9221$ in the second case.

%

\section{Experimental methodology}
\label{sec.exp_method}

In this section, we describe in detail the backbone architecture of the network, the procedure of dataset construction and the training setting considered in our experiments.

The
proposed solution is compared with the  state-of-the-art approach in  \cite{niu2019SPL} based on deep learning and, for the aligned case, also with those in \cite{Galvan2014,Bianchi2012} based on
statistical analysis. While in fact the method in  \cite{niu2019SPL} always
outperforms all the other previous methods for the non-aligned scenario (e.g. \cite{Bianchi2012,Dalmia2018}), for the aligned case,  there are some cases where the accuracies achieved by the methods in \cite{Galvan2014,Bianchi2012}, tailored for the aligned scenario, are superior to those of \cite{niu2019SPL}.

%


\subsection{Backbone architecture}
\label{sec.architecture}

For the design of the internal layers of the CNN, we considered the DenseNet architecture \cite{huang2017densely},  which was also considered in \cite{niu2019SPL}.
Such backbone architecture has been
recently adopted for several image forensic tasks, see for instance
\cite{chen2019multi,kamal2018applicationDense,yang2014effective}, yielding improved
performance compared to those achieved with traditional CNN architecture (e.g.
residual-based networks).

%
%
The main feature of the dense structure is that it connects each layer to every other layer in a dense block in a feed-forward fashion. in this way, the features extracted by the various
layers are used by the subsequent layers throughout the same
dense block (hierarchical structure).
The dense connectivity has been shown in \cite{huang2017densely} to mitigate the  gradient vanishing problem.
%
The number of links in the network increases compared to traditional CNN architectures, passing from $l$ to $l(l-1)/2$ for each dense block, where $l$ is the number of layers in the block.
However, as an advantage, the number of (to-be-trained) parameters is significantly reduced.
%
%
Following the original dense structure (see \cite{huang2017densely}), we considered a network depth of $40$, with 3 dense blocks and growth rate $k = 12$.
Each dense layer consists of 12 convolutional layers and a transition layer, where $2\times 2$ average pooling is performed to decrease the input size.  All the convolutions have kernel size $3 \times 3 \times 12$.
The default dropout of $0.2$ is considered.
An initial convolution with $24$ ($2 k$)  filters of size $3 \times 3$ is performed before the first dense block.
For more details on the dense structure we refer to \cite{huang2017densely}.
After the last dense blocks, global average pooling is performed and the feature vector is fed to the fully connected layer.

The number of  output nodes of the fully-connected layer is set  to $(q_{1,1}^M \cdot q_{1,2}^M \cdot \cdots \cdot q_{1,N_c}^M)$. A softmax is applied to each block of $q_{1,i}^M$ nodes independently, for a total of $N_c$ softmaxes, as illustrated in Figure \ref{fig:setup}.

\subsection{Datasets} 

As in \cite{niu2019SPL}, a model for $Q_1$ estimation is trained for a fixed value of $QF_2$. This does not represent a limitation in practice since the information on the second compression is always available. The knowledge of the final quantization matrix, in fact, can be recovered from the JPEG file, and is necessary to decompress the image, getting the image in the pixel domain. Moreover, when the image is re-saved in an uncompressed format, the quantization matrix of last compression can be accurately estimated \cite{bestagini2012video}.
As a drawback,  a model has to be trained for every matrix of second quantization, which may be time-consuming (the same happens with the method in \cite{niu2019SPL}). However, since training has to be performed only once, this does not represent a big issue. Moreover, our experiments show that a network trained for a given $QF_2$ generalizes pretty well to a different $QF_2$'s or, more in general, to a different $Q_2$ matrix, when the difference is not too much (a $\pm$ 2 mismatch in the QF resulting in a very small decrease of performance), hence a limited number of models can be trained.
The training and testing datasets are built as described in the following.

We considered the RAISE dataset \cite{RAISE8K}  with 8156
native (tiff) images, that is split into a training and a test set. Specifically,  7000 images were considered for training, while the remaining 1156 were reserved for testing.
The images were then compressed first with several $QF_1$'s and then with the prescribed $QF_2$, thus obtaining several double compressed versions (both $QF_1$ values larger and smaller than $QF_2$ were considered).
JPEG compression was performed with OpenCV.
%
To simulate the misalignment,  we applied a random grid shift $(r,c)$ with $0 \le r, c \le 7$ between the two compressions, with $r, c $ randomly selected in the $[0:7]$ range. Therefore, the JPEG is non-aligned with probability 63/64, while the aligned scenario (which corresponds to the case $r = c = 0$) occurs with probability 1/64.

To build the dataset of patches used for training, we proceeded as follows: for every $QF_1$, we cropped the DJPEG images in the training set into patches of size $64\times 64\times 3$; then,  from each image  we took 100 patches in random positions;
we stopped collecting patches when a total number of $10^{5}$ patches
was reached (coming then from 1000 images) for each given $QF_1$.\footnote{For every $QF_1$, a random shuffle was applied to the 7000 DJPEG images compressed with $(QF_1, QF_2)$, so the subset of images considered for every $QF_1$ was never the same.}
%
%
%
%
For our experiments we set $QF_2=90$ and $80$.
Specifically, for $QF_2=90$, we built the training dataset ${\mathcal{D}}^{90}$ by considering $QF_1 \in [60:98]$, for a total of  $3.9 \times 10^{6}$ patches.
For $QF_2=80$, we built set ${\mathcal{D}}^{80}$ by considering $QF_1 \in [55:98]$, then for a total of  $4.4 \times 10^{6}$ patches.
The test patch set was obtained in the same way. In this case, for every $QF_1$, all the 1156 images in the test set were considered, each one contributing to 100 random patches (for a total of 115600 patches).
\CH{By following \cite{niu2019SPL}, we directly feed the 3 ($R$, $G$ and $B$) channels to the CNN.
%
Another possibility would be to first perform color space conversion from $R G B$ to  $Y C_b C_r$ and then feed the transformed image to the CNN (since passing from $R G B$ to $Y C_b C_r$ corresponds to the application linear mapping, the network should in principle be able to learn the mapping itself, if it benefits the learning process).}

The Dresden dataset \cite{Dresden} was also considered to test the performance under dataset mismatch, consisting of 1491 raw images (hence, for each $QF_1$, the performance were tested on a total of 149100 patches).
\CH{The size of the images is around 3600$\times$2700, which is a bit smaller than the size of RAISE images (4928 $\times$ 3264).
To further investigate the behavior of our method in presence of  a resolution mismatch, we also consider subsampled versions of the raw images from the Dresden database, corresponding to a resolution less than half of the resolution considered for training.}

\CH{
%
Finally, we also run some tests in the case where the first compression is carried out by using Photoshop (PS),
that does not use
standard quantization matrices, thus representing a case of strong
mismatch between training and testing.}


%
%
%

\subsection{Setting}

In all the experiments, we set $N_c = 15$,
which is the value considered in most prior works \cite{Bianchi2012,Cogranne2019,Galvan2014,Dalmia2018, niu2019SPL}. Hence, we estimate the first 15 DCT coefficients, zig-zag scanned.
%

For the implementation of the proposed method and the custom loss, we used TensorFlow version 2.2.
Model training and
testing were carried out in Python via TensorFlow, using Keras API.

We ran our experiments using
a 2x Nvidia GeForce RTX 2080 Ti 11 GB GDDR6 GPU.
For the optimization, the Adam solver was used with learning rate $10^{-5}$.
The batch size for training and testing was set to 32 images.
We got our models using the $\mathcal{L}^r$ loss by training the network for 100 epochs.
After this number of epochs we verified that the loss decreases very slightly (less than 0.01\% at every iterations) and the accuracy of the estimation cannot be improved further by letting the training go on (only incurring the risk of overfitting).
The weight in the combined loss $\mathcal{L}^r$ in Eq. \eqref{eq.loss_comb} was set to $c=0.8$.


%
%
The code is publicly available and can be found at the github link
\url{https://tinyurl.com/yxhl32w5}.


\begin{figure}[th!]
\begin{center}
\includegraphics{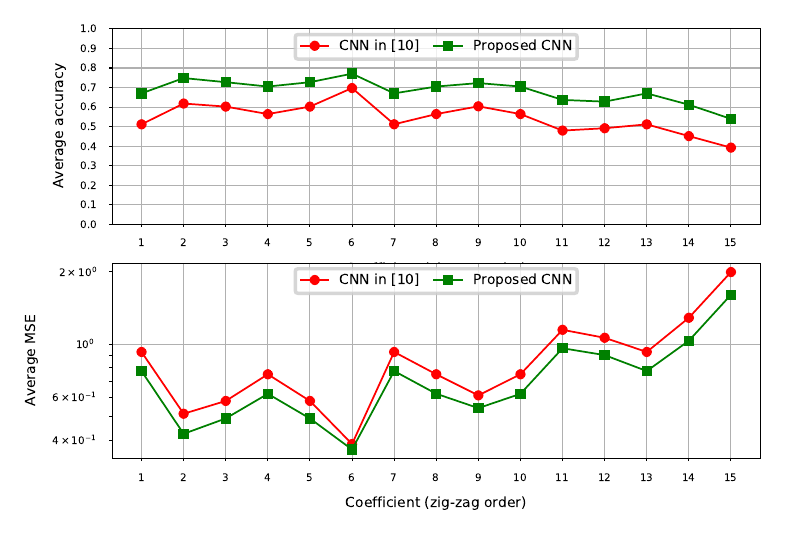}
\caption{Average accuracy (top) and average MSE (bottom) of the estimation for each of the 15 DCT coefficients, for $QF_2 = 90$, in the non-aligned DJPEG scenario. 
}
\label{fig:AccuracyPerf_DCTcoeff}
\end{center}
\end{figure}
\begin{figure}[th!]
\begin{center}
\includegraphics{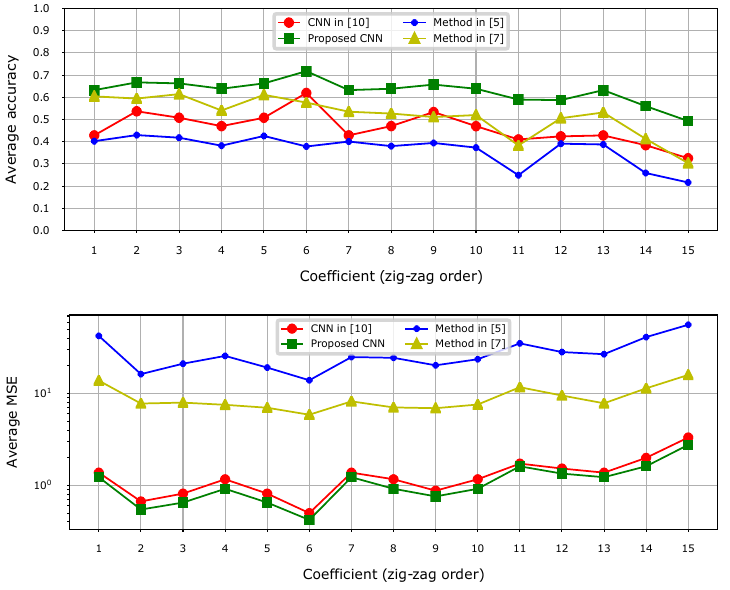}
\caption{Average accuracy (top) and average MSE (bottom) of the estimation for each of the 15 DCT coefficients, for $QF_2 = 90$, in aligned DJPEG scenario.
}
\label{fig:MSEPerf_DCTcoeff}
\end{center}
\end{figure}

\begin{figure}[th!]
\begin{center}
\includegraphics{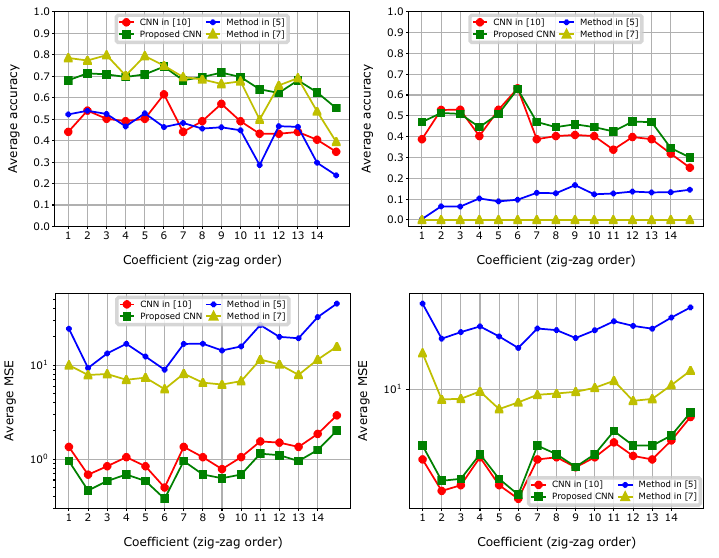}
\caption{Average accuracy when $QF_1 < QF_2$ (top left) and when $QF_1 \ge QF_2$ (top right), average MSE when $QF_1 < QF_2$ (bottom left) and when $QF_1 \ge QF_2$ (bottom right) of the estimation for each of the 15 DCT coefficients, for $QF_2 = 90$, in aligned DJPEG scenario.}
\label{fig:seperate_DCTcoeff}
\end{center}
\end{figure}

\section{Results}
\label{sec.results}

\subsection{Comparison with existing methods}

The average performance results achieved by the CNN model for the new architecture, trained with the $\mathcal{L}^r$  loss on ${\mathcal{D}}^{90}$, are reported in Table \ref{tab:tableResStep1_avg},
where they are compared to those achieved by the CNN model in \cite{niu2019SPL}, trained for the same value of $QF_2 = 90$.
The performance results are averaged under the same setting considered for the training  regarding the alignment of the DJPEG, i.e. the test patches are  DJPEG,    aligned with probability 1/64, non-aligned   with 63/64.
%
The performance results of the new model are superior to those achieved by the method in \cite{niu2019SPL} both in terms of accuracy and of MSE.
\begin{table}[t!]
   \begin{tabular}{l|c|c}
    \hline
    & Prop CNN & CNN in \cite{niu2019SPL} \\ \hline
    AvgAcc & {\bf 0.689}  & 0.547  \\
    AvgMSE  & {\bf 0.731} & 0.882
\end{tabular}%
\vspace{0.2cm}
\caption{Average performance of the proposed CNN estimator and \cite{niu2019SPL} for $QF_2 = 90$. The DJPEG is non-aligned with probability 63/64, aligned with probability 1/64 (same setting considered for the training of the models).}
\label{tab:tableResStep1_avg}
\end{table}
%
%


\begin{table}[t!]

   \begin{tabular}{l|c|c|c|c}
    \hline
    & Prop CNN & CNN in \cite{niu2019SPL} & \cite{Bianchi2012} & \cite{Galvan2014}\\ \hline
    AvgAcc & {\bf 0.627}  & 0.463 & 0.366 & 0.518 \\
    AvgMSE  & {\bf 1.120} & 1.326  & 28.016 & 9.091
\end{tabular}%
\vspace{0.2cm}
\caption{Average performance of the proposed CNN estimator in the aligned case for $QF_2 = 90$, and comparison with the state-of-the-art.}
\label{tab:tableResStep1_avg_Al}
\end{table}
The performance results achieved by the methods in the aligned DJPEG scenario are reported in Table \ref{tab:tableResStep1_avg_Al}.
The performance results in aligned scenario are slightly inferior to those reported in the Table \ref{tab:tableResStep1_avg} for the mixed case.
From the results of Table \ref{tab:tableResStep1_avg_Al},  we see that the proposed method can also outperform \cite{Galvan2014}, which is specifically designed for the aligned case, both in terms of MSE and accuracy.

%
%

The estimation accuracy and MSE for each DCT coefficient are reported for the non-aligned and aligned case in Figure \ref{fig:AccuracyPerf_DCTcoeff} and \ref{fig:MSEPerf_DCTcoeff}, where the results are averaged on all the $QF_1$s.
Comparison with the methods \cite{Galvan2014,Bianchi2012} is also reported for the aligned case.
It can be observed that the proposed method greatly outperforms these methods,
for all the 15 DCT coefficients.
%

Figure \ref{fig:seperate_DCTcoeff} shows the averaged results when $QF_1 < QF_2$ and $QF_1 \ge QF_2$ respectively, in the case $QF_2$=90 for the aligned scenario. It can be noticed that when $QF_1 \ge QF_2$ the proposed method clearly outperforms the methods \cite{Galvan2014,Bianchi2012}, both in terms of accuracy and MSE. When $QF_1 < QF_2$, the proposed method still outperforms the state-of-the-art methods.

\begin{figure}[th!]
\begin{center}
\includegraphics{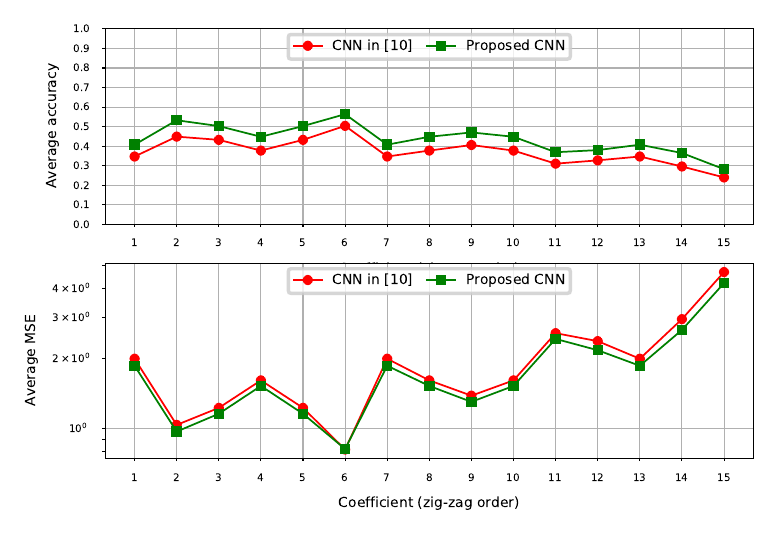}
\caption{Average accuracy (top) and MSE (bottom) of the estimation for each of the 15 DCT coefficients, for $QF_2 = 80$, in the non-aligned DJPEG scenario.}
\label{fig:AccuracyPerf_DCTcoeff80}
\end{center}
\end{figure}

\begin{figure}[th!]
\begin{center}
\includegraphics{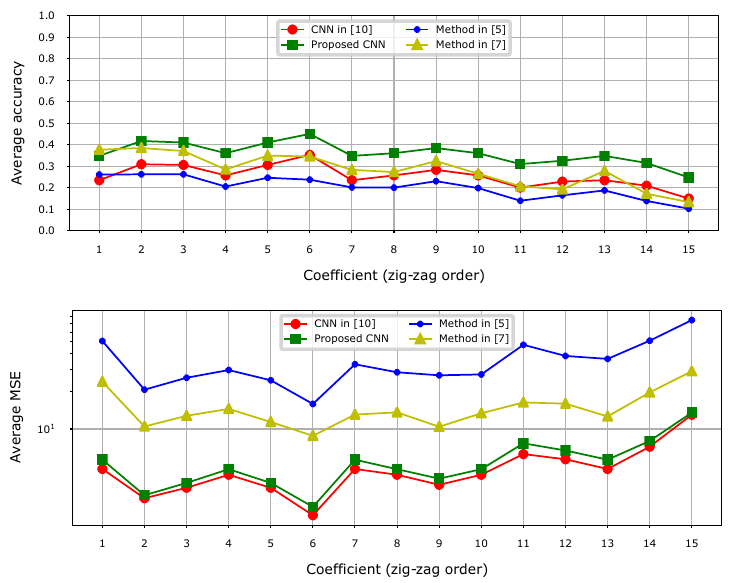}
\caption{Average accuracy (top) and  MSE (bottom) of the estimation for each of the 15 DCT coefficients, for $QF_2 = 80$, in the aligned DJPEG scenario.}
\label{fig:MSEPerf_DCTcoeff80}
\end{center}
\end{figure}

The average performance obtained for the case $QF_2 = 80$ (model trained on ${\mathcal{D}}^{80}$) are reported in Table \ref{tab:tableResStep1_avg80} for the non-aligned scenario and Table \ref{tab:tableResStep1_avg80_Al} for the aligned scenario.
A performance loss is experienced by all the methods. This was expected since with a smaller $QF_2$, the second
quantization tends to erase more the traces of the first compression, thus making the estimation harder.
Nevertheless, the proposed method has an advantage over the
state-of-the-art.
We see that the CNN model in \cite{niu2019SPL} does better than our method  in terms of MSE for the aligned case; however, our method outperforms \cite{niu2019SPL}  in terms of accuracy in the aligned case and both in  terms of accuracy and MSE in the non-aligned case, which is our main focus.
Figure \ref{fig:AccuracyPerf_DCTcoeff80} and \ref{fig:MSEPerf_DCTcoeff80}  report the results on 15 DCT coefficients
in the case $QF_2 = 80$, averaged on all the $QF_1$s, for the non-aligned and aligned case respectively.
We see that the gain of the proposed method is confirmed.

\subsection{Generalization capability}
The generalization capability of the model are tested by considering several sources of mismatch, i.e.,  the second  compression quality $QF_2$, and the image database.

\begin{figure}[t!]
\begin{center}
\includegraphics{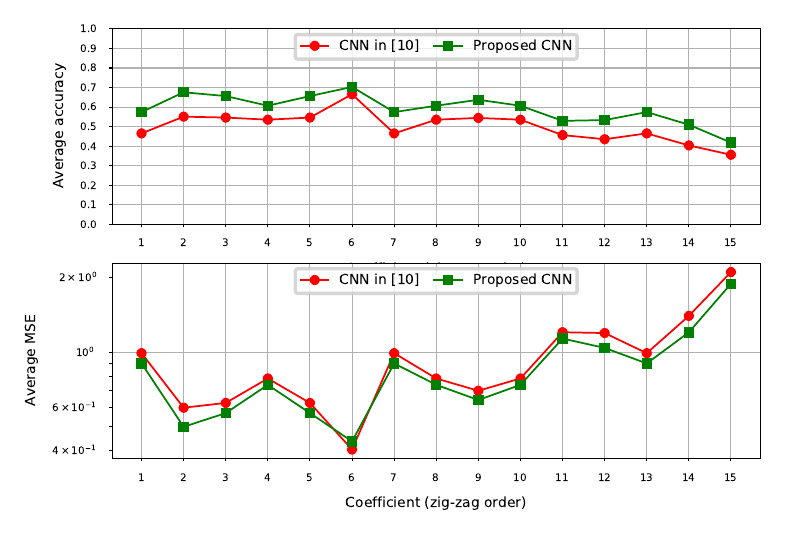}
\caption{Performance of the CNN estimator under mismatched $QF_2$ ($QF_2 = 92$). Average accuracy (top) and MSE (bottom) of the estimation for each of the 15 DCT coefficients. }
\label{fig:Q2mismatch}
\end{center}
\end{figure}

The results in presence of $QF_2$ mismatch are reported in
Figure \ref{fig:Q2mismatch}, where the model trained on ${\mathcal{D}}^{90}$ is tested on images compressed with  $QF_2 = 92$, in the same general setting regarding the alignment considered for training (that is, the DJPEG is aligned with probability 1/64). We see that,
the drop of performance is limited, proving a certain generalization capability, and is similar for the two methods. The performance of the proposed CNN remains superior to \cite{niu2019SPL}: the total AvgAcc and AvgMSE are respectively 0.591 and 0.859 for our method, and 0.500 and 0.944   for the method in \cite{niu2019SPL}.
\CH{Notably, in the aligned scenario, the performance remains superior to those of the state-of-the art methods in  \cite{Galvan2014} and \cite{Bianchi2012} designed for this case. Specifically, the  AvgAcc and AvgMSE are  0.579   and  1.029 for our method,  0.331 and 15.8 for  \cite{Galvan2014}, and 0.397 and 35.5 for  \cite{Bianchi2012}.}

\begin{figure}[t!]
\begin{center}
\includegraphics{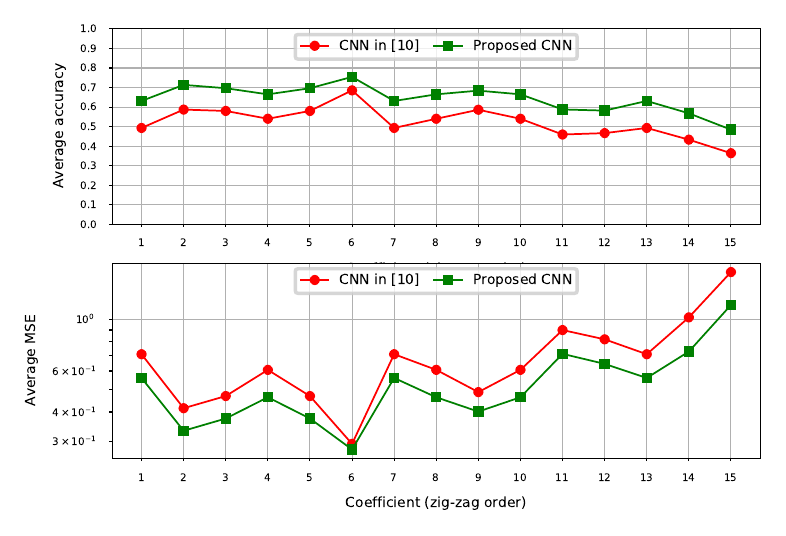}
\caption{Performance of the estimator on DJPEG images from a different database (Dresden), for $QF_2 = 90$. Average accuracy (top) and MSE (bottom) of the estimation for each of the 15 DCT coefficients.}
\label{fig:DBmismatch}
\end{center}
\end{figure}

\begin{figure}[t!]
\begin{minipage}{0.95\linewidth}
	\begin{center}
		\includegraphics[width = 8cm]{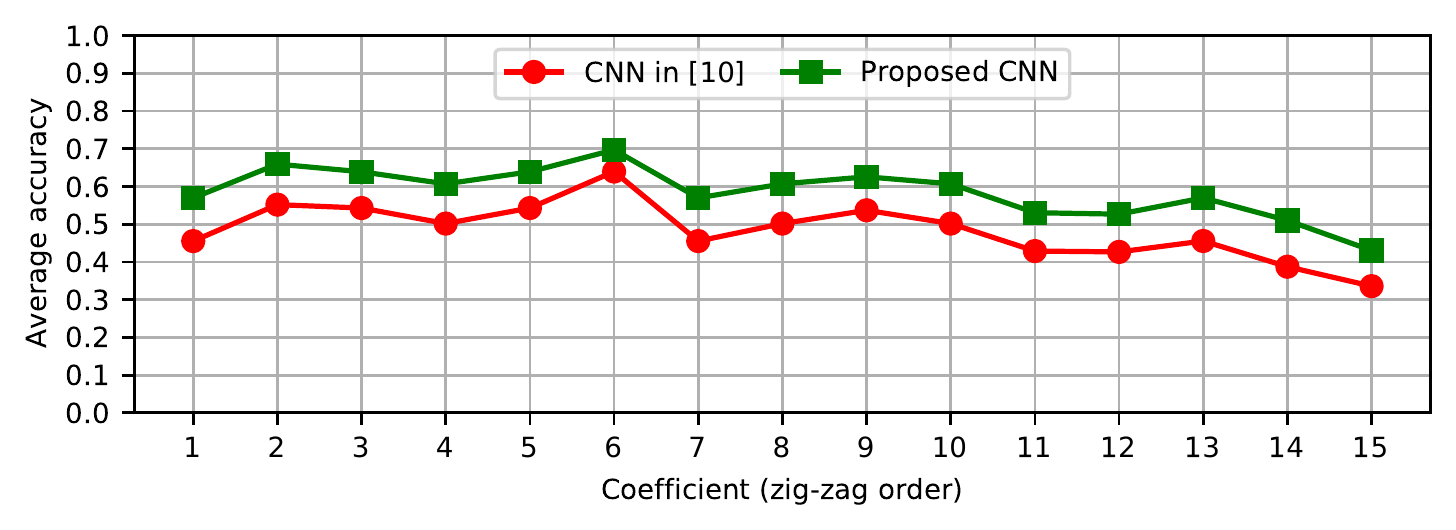}
	\end{center}
\end{minipage}

\begin{minipage}{0.95\linewidth}
	\begin{center}
		\includegraphics[width = 8cm]{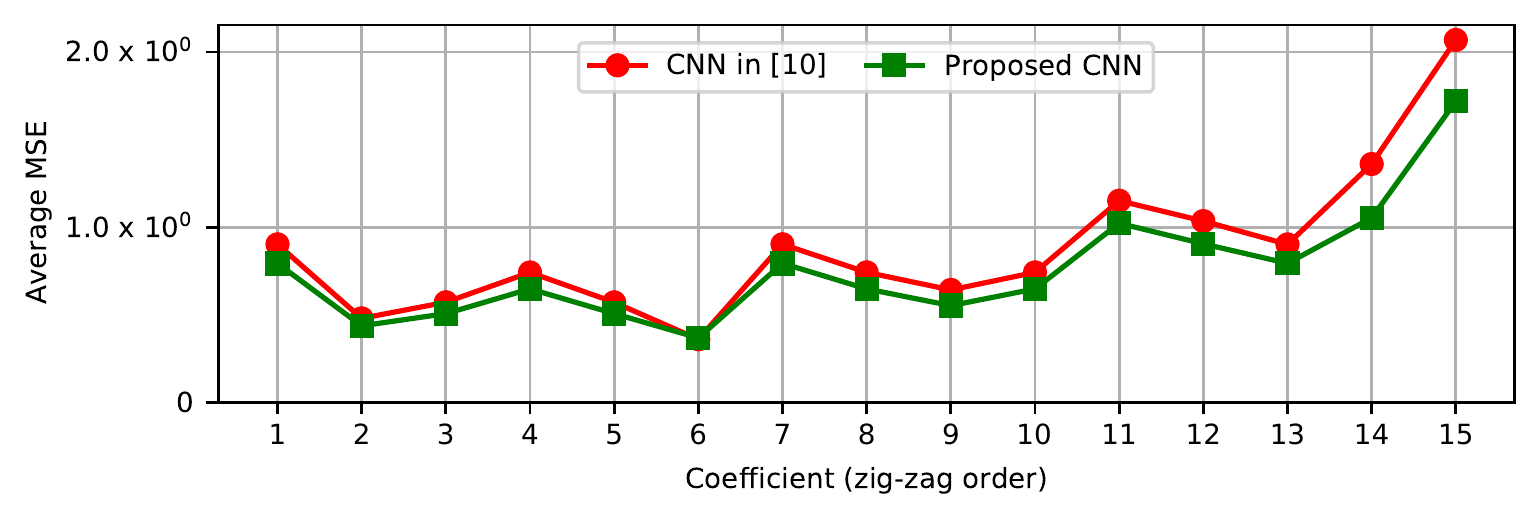}
	\end{center}
\end{minipage}
\caption{\CH{Performance of the estimator on DJPEG low resolution images from Dresden dataset ($QF_2 = 90$). Average accuracy (top) and MSE (bottom) of the estimation for each of the 15 DCT coefficients. }}
\label{fig:DBmismatch-low}

\end{figure}

To assess the impact that dataset mismatch has on the performance of our CNN-based estimator, we also evaluate the performance of the estimator on DJPEG images coming from the Dresden dataset.
Figure \ref{fig:DBmismatch} reports the results of our tests for $QF_2 = 90$.
{The total AvgAcc and AvgMSE are respectively 0.644 and 0.538 for our method, and 0.523 and 0.694 for the method in \cite{niu2019SPL}.}
%
\CH{The results with the half-resolution images from Dresden dataset (strong resolution mismatch) are reported in Figure \ref{fig:DBmismatch-low}. As expected, the performance decreases, but not seriously so, and the superior performance of our method over \cite{niu2019SPL} are confirmed also in this case.}


%
%


\begin{table}[th!]

   \begin{tabular}{l|c|c}
    \hline
    & Prop CNN & CNN in \cite{niu2019SPL} \\ \hline
    AvgAcc & {\bf 0.440}  & 0.375 \\
    AvgMSE  & {\bf 1.866} & 1.946
\end{tabular}%
\vspace{0.2cm}
\caption{Average performance of the proposed CNN estimator and \cite{niu2019SPL} for $QF_2 = 80$ . The DJPEG is non-aligned with probability 63/64 (same setting considered for the training of the models), aligned with probability 1/64.}
\label{tab:tableResStep1_avg80}
\end{table}
\begin{table}[th!]

   \begin{tabular}{l|c|c|c|c}
    \hline
    & Prop CNN & CNN in \cite{niu2019SPL} & \cite{Bianchi2012} & \cite{Galvan2014}\\ \hline
    AvgAcc & {\bf 0.360}  & 0.254 & 0.202 & 0.282 \\
    AvgMSE  & 5.594 & {\bf 4.970}  & 35.453 & 15.061
\end{tabular}%
\vspace{0.2cm}
\caption{Average performance of the proposed CNN estimator in the aligned case for $QF_2 = 80$, and comparison with the state-of-the-art.}
\label{tab:tableResStep1_avg80_Al}
\end{table}

\begin{figure}[t!]
\begin{center}
\includegraphics{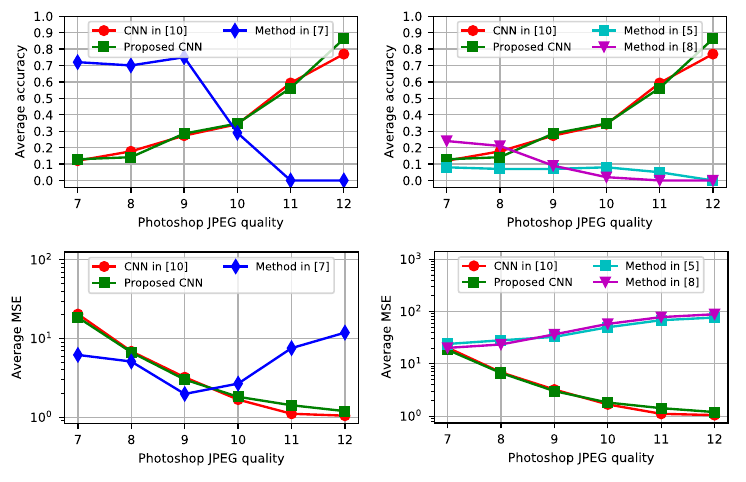}
\caption{\CH{Performance of the estimator when the first compression is performed with Photoshop for several PS qualities ($QF_2 = 90$), in the aligned (first column) and non-aligned (second column) scenario.
}.}
\label{fig:PSqualities}
\end{center}
\end{figure}

\CH{Finally, the results of our tests in the case where the first compression is performed with non-standard quantization matrices are provided in Figure \ref{fig:PSqualities}, where we report the accuracy and MSE achieved for some common medium-high Photoshop qualities (compression levels from 7 to 12), respectively for the aligned and non-aligned case. The performance are compared with \cite{niu2019SPL} and  with the baseline model-based methods for this case. 
The performance of \cite{niu2019SPL} and the proposed methods are very similar, our method being superior only slightly on the average  and in terms of MSE. For the non-aligned case, both methods significantly outperform the method in  \cite{Bianchi2012} and the one in \cite{Dalmia2018} for PS qualities above 8.
%
In the aligned case, instead,
the tailored method in \cite{Galvan2014} works better for low PS qualities, while,
for higher qualities, the CNN-based
estimators gets much better performance (the case where
the first quantization step is smaller than the second one is  a
scenario that model-based techniques can not handle properly).
Clearly, the performance of our CNN estimator for low PS qualities can be improved if the model is trained, or even just fine-tuned, considering examples of JPEG images compressed with non-standard  quantization matrices.
%
%
}

\subsection{Application to tampering localization}
\label{sec.localization}

Given the capability of our CNN  estimator to work on small image patches, quite straightforwardly,  the method can be exploited to localize possible tampering regions in a DJPEG image. Specifically, given a DJPEG tampered image, the estimator can applied  on sliding windows   to get a map with the estimated primary quantization coefficients $(\hat{{\bf q}_1})_{N_c}$ for each $64\times 64$ block. 
Notably,
by looking at those maps  the tampering can be exposed
in the general scenario where both the background and the foreground (tampered areas) are DJPEG, that is, both the background and the copy-pasted region were originally JPEG (compressed with a different qualities) and undergo a second JPEG compression after forging. This corresponds to a very common scenario in practice. In this scenario, methods that try to detect and localize tampering by looking at the presence or absence  of typical double compression artifacts do not work, see for instance \cite{bianchi2011improved,amerini2014splicing,wang2016double,Barni2017}, just to mention a few of them.
These methods in fact  implicitly assume that the background  is single  compressed, while the foreground is double  compressed, or viceversa.\footnote{\CH{As we mentioned in the introduction, for a single compressed region, our estimator returns a quantization matrix of a very high compression  quality  ( to give an insight,
the coefficients of the quantization matrix for QF=100 are estimated with an accuracy larger than 0.8). Therefore, when the method is applied for tampering localization in this scenario, the manipulation can still be exposed based on the inconsistencies between the estimated coefficients of background and foreground. }
}

In order to get a localization map, we first divide the input image ${\bf x}$ of size $V \times L \times 3$ into overlapping blocks of size 64 $\times$ 64 with stride $s = 1$; then, each block is fed to the CNN that returns a vector with the first $N_c$ estimated quantization coefficients.
Let \mbox{$QM(i,j,:) = f([{\bf x}_{ij}])$} be the network output when the input is
the $(i,j)$-th image block of size 64 $\times$ 64 $\times$  3; then, $QM(i,j,:) = ({\hat{\bf q}}_1([{\bf x}_{ij}]))_{N_c}$. In this way, for each $k = 1,...,N_c$ we obtain a map $QM(:,:,k)$ with the estimated values of the $k$-th coefficient for each block.

  \begin{figure*}[h!]
	\includegraphics{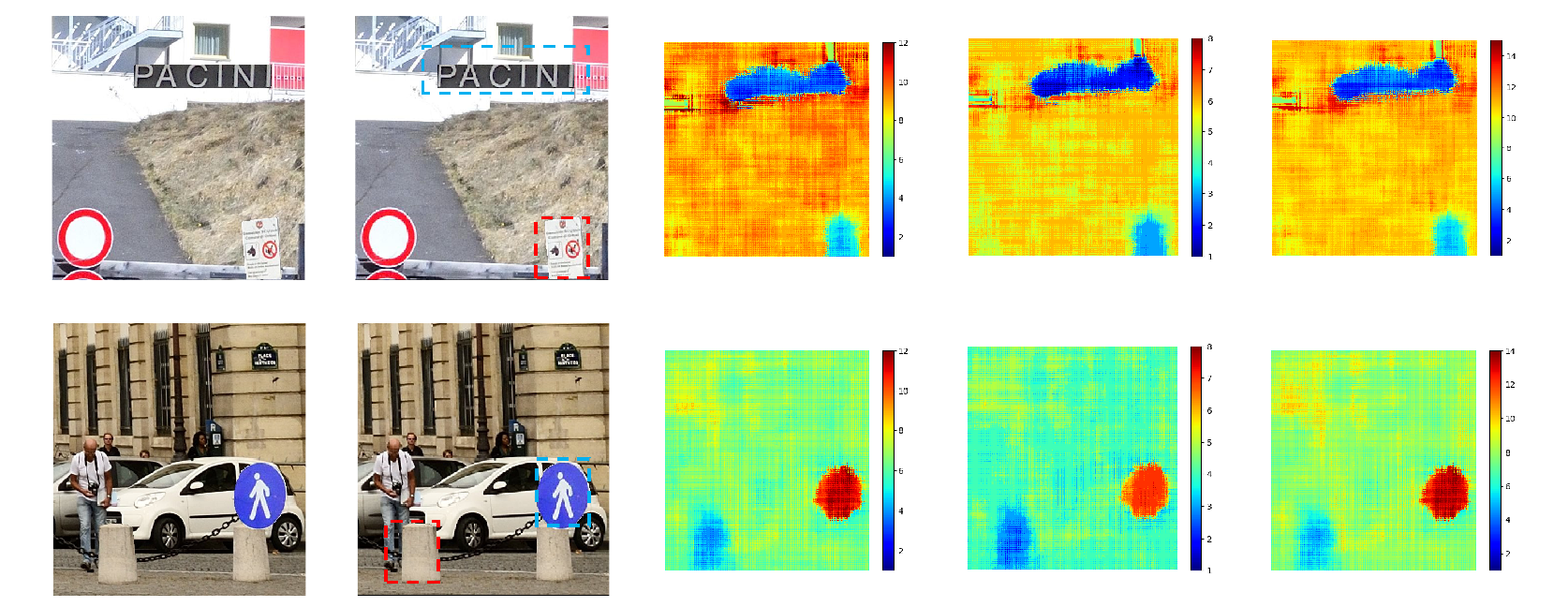}
	\caption{ 
		Examples of tampered (double JPEG) images and  map of the estimated $k$-th $Q_1$ coefficient, for some values of $k$.
		{For each example, we report, from left to right: the tampered image, the ground-truth tampering map,  the estimated map for the DCT coefficient no 1, 6 and 14, that is  $QM(:,:,1)$,  $QM(:,:,6)$ and  $QM(:,:,14)$}.
	}
	\label{tampered-imgs}
\end{figure*}

We performed a qualitative analysis. Figure \ref{tampered-imgs} shows two examples. {For both tampered images, we have two distinct tampered areas, where the copy-pasted regions have different first JPEG qualities, that is, $QF_{1,1}$ = 95 and $QF_{1,2}$ = 85 for the first example and $QF_{1,1}$ = 65 and $QF_{1,2}$ = 95 for the second one. The first JPEG quality for the background of the two examples is 75. The last quality factor for both examples is $QF_2 = 90$. All the JPEG grids are not aligned.}
{For sake of visualization a color map is reported.
The color map shows that the two tampering regions have a  different $q_{1,k}$ from the background;
interestingly, the color map also reveals  that the  $q_{1,k}$ value is also different between them, hence, that they correspond to two distinct tampering (the copy-pasted regions come from different donor images, having different JPEG compression qualities).}

In general, if the qualities of the former JPEG are close, it is harder to visualize and expose the tampering by simple inspection of the $N_c$ maps of the $q_{1,k}$ coefficients of each block, that is $QM(:,:,k)$, all the more that some of the coefficients might have the same value. In this cases, we could resort to clustering  to get a tampering localization map from the vectors of the estimated $q_{1,k}$ values for each position. This interesting analysis is left as a future work.

\section{Conclusions}
\label{sec.conclusions}

In this paper, we proposed a method for primary quantization matrix estimation via CNNs, that resorts to a classification-like architecture to perform the estimation of the quantization coefficients.
Thanks to the adoption of a simil-classification structure, the new CNN estimator achieves improved performance with respect to the CNN regression-based method in \cite{niu2019SPL}, both in terms of accuracy and MSE.


Notably, the proposed method is a general one, which can work under a wide variety of operative conditions.
i.e. when the second compression grid is
either aligned or not with the one of first compression, and for every combinations of qualities of former
and second compression.
Regarding the JPEG alignment, the method is designed to work in particular for the case of non-aligned double JPEG compression (the
aligned case is assumed to occur with probability 1/64).
A method capable to deal with primary quantization estimation in the non-aligned scenario, in fact, is very relevant when the proposed estimator is used for image tampering localization
(when a region of a JPEG image is copy-pasted
into another JPEG image, in fact, very likely, the alignment
between the compression grids is not preserved and the final
JPEG is non aligned with the grid of the spliced area).
%
%
%
Despite its generality, the proposed method also outperforms the existing - dedicated - state-of-the-art solution for the aligned scenario in most of the cases.
The method provide  very good performance also in the challenging case of $QF_1 > QF_2$, where state-of-the-art methods based on  statistical analysis often fail. More importantly, the estimator works on small image patches, that opens the way to the application of the method for tampering localization  (see Section \ref{sec.localization}).

As future research,  the application of the method for tampering localization in DJPEG images,  possibly including the identification of the different tampering sources (donors)
is  an interesting direction. 
In addition, the robustness of the estimator in presence of adversarial attacks could also be investigated.
From a more general perspective, we believe that a similar architecture to the one proposed in this paper could also be exploited to address other estimation problems which are relevant in image forensics.

\begin{backmatter}

\section*{Funding}

%
%

This work has been partially supported by the National Natural Science
Foundation of China (Grant 61872244), Guangdong Basic and Applied
Basic Research Foundation (Grant 2019B151502001), Shenzhen R\&D Program
(Grant JCYJ20200109105008228, JCYJ20180305124325555)
and by the Italian Ministry of University and Research (MUR) under the PRIN 2017
2017Z595XS-001 program - PREMIER project.

\section*{Abbreviations.}

JPEG: Joint Photographic Experts Group. DJPEG: double JPEG. DCT: dIscrete cosine transform. QF: quality factor. CNN: convolutional neural network. MSE: mean squared error. AvgACC: average accuracy. AvgMSE: average mean squared error. API: application programming interface.

\section*{Availability of data and materials.}

The datasets used for the experiments are publicly available in \cite{RAISE8K} and \cite{Dresden}.
The code is available at the following Github link: \url{https://github.com/andreacos/BoostingCNN-Jpeg-Primary-Quantization-Matrix-Estimation}.

\section*{Competing interests}
  The authors declare that they have no competing interests.

\section*{Author's contributions}

BT conceived the presented idea and developed the theoretical formalism.
BL supervised the study and helped interpreting the results.
BT wrote the paper, with the support of BL.
AC and BT conceived and planned the experiments.
AC implemented the method and helped interpreting the results.
DH and AC carried out the experiments and positioned the work in the current state of research.




\bibliographystyle{bmc-mathphys} 
\bibliography{Qestimation}      




\end{backmatter}
\end{document}